\documentclass[letterpaper]{article} 

\usepackage{aaai23}  
\usepackage{times}  
\usepackage{helvet}  
\usepackage{courier}  
\usepackage[hyphens]{url}  
\usepackage{graphicx} 
\usepackage{natbib}  
\usepackage{caption} 
\usepackage{xcolor}
\usepackage{amsmath}
\usepackage{amsfonts}
\usepackage{bm}
\usepackage{amssymb}
\usepackage{graphicx}
\usepackage{subcaption}
\usepackage{caption}
\usepackage{multicol}
\usepackage{multirow}

\usepackage[boxed,linesnumbered,ruled]{algorithm2e}

\graphicspath{{figures/}}

\newlength\alignwidth

\usepackage{newfloat}

\makeatletter
\newcommand{\removelatexerror}{\let\@latex@error\@gobble}
\makeatother

\title{Understanding and Enhancing Robustness of Concept-based Models}
\author{
    Sanchit Sinha\textsuperscript{\rm 1}, Mengdi Huai\textsuperscript{\rm 1,2}, Jianhui Sun\textsuperscript{\rm 1}, Aidong Zhang\textsuperscript{\rm 1}\\
}
\affiliations{
    \textsuperscript{\rm 1}University of Virginia\\
    \textsuperscript{\rm 2}Iowa State University\\

    \textsuperscript{\rm 1}\{sanchit, js9gu, aidong\}@virginia.edu, \textsuperscript{\rm 2}mdhuai@iastate.edu
}

\begin{document}
\maketitle

\begin{abstract}
Rising usage of deep neural networks to perform decision making in critical applications like medical diagnosis and financial analysis have raised concerns regarding their reliability and trustworthiness. As automated systems become more mainstream, it is important their decisions be transparent, reliable and understandable by humans for better trust and confidence. To this effect, concept-based models such as Concept Bottleneck Models (CBMs) and Self-Explaining Neural Networks (SENN) have been proposed which constrain the latent space of a model to represent high level concepts easily understood by domain experts in the field. Although concept-based models promise a good approach to both increasing explainability and reliability, it is yet to be shown if they demonstrate robustness and output consistent concepts under systematic perturbations to their inputs. To better understand performance of concept-based models on curated malicious samples, in this paper, we aim to study their robustness to adversarial perturbations, which are also known as the imperceptible changes to the input data that are crafted by an attacker to fool a well-learned concept-based model. Specifically, we first propose and analyze different malicious attacks to evaluate the security vulnerability of concept based models. Subsequently, we propose a potential general adversarial training-based defense mechanism to increase robustness of these systems to the proposed malicious attacks. Extensive experiments on one synthetic and two real-world datasets demonstrate the effectiveness of the proposed attacks and the defense approach.
\end{abstract}

\section{Introduction}
\begin{figure}[h]
    \includegraphics[width=0.47\textwidth]{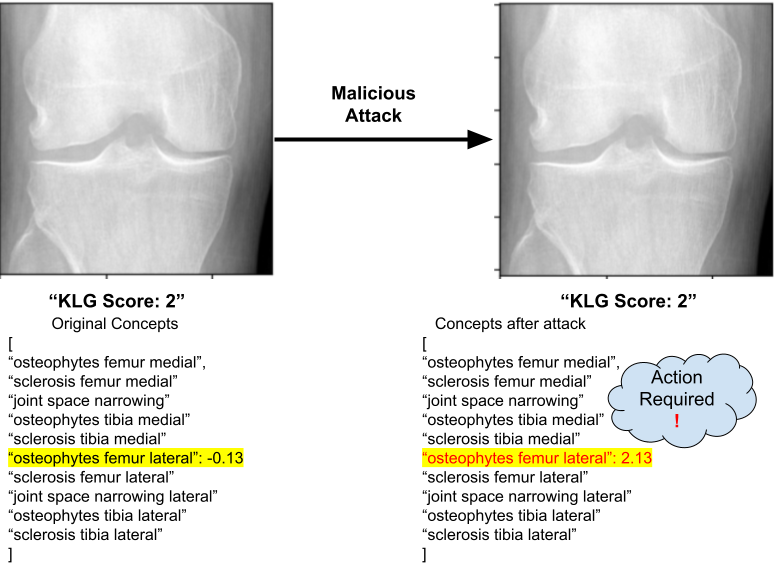}
\vskip -8pt
    \caption{Example of how a value of a concept indicating osteophytes (bone spurs) can be maliciously changed although the actual severity of disease quantified by Kellgren-Lawrence grade (KLG Score) remains the same.}
    \label{fig:example-attack-oai}
 \vskip -10pt
 \end{figure}

With growth of highly specialized architectures for a variety of use-cases and their superior performance, Deep Neural Networks (DNNs) are increasingly being used in sensitive and critical applications such as medical diagnosis,  employment/recruiting, financial credit analysis, etc. However, widespread adoption of such models faces several challenges - primarily the black-box nature of their predictions. Many recent research works have proposed explanation methods which provide deeper understanding of model predictions by providing ``explanations''. Explanations range from being local in nature where they assign importance scores to the features present in an input sample but can also be global in nature where the model identifies certain ``concepts'' present in the input sample. A concept can be thought of as an abstraction of features which are usually shared across multiple similar sample points. For example, in Figure~\ref{fig:example-attack-oai}, a concept can be entirely clinical ``osteophyptes- femur'', ``sclerosis-tibia'', etc. Usually DNNs are trained end-to-end, which makes it difficult to isolate concepts and even harder to make them human understandable. To alleviate this, concept-based approaches have been proposed \cite{koh2020concept,alvarez2018towards} which map a sample from input space to a concept space and subsequently map the concept space to the prediction space. The concept space usually consists of high-level human understandable concepts. A model trained incorporating either manually curated or automatically learned concepts increases both interpretability and reliability of its predictions. One such example, as proposed in \cite{koh2020concept}, Concept Bottleneck Models (CBMs), can help domain experts quickly identify any discrepancy and intervene when and where needed. CBMs also offer generalizability in the sense that any DNN can be easily converted into a CBM by resizing an intermediate layer to correspond to the size of any closed concept set pre-selected by domain experts. The training of such models uses standard training procedure with a loss function augmented with an extra term from the bottleneck layer.

Although incredibly simple in formulation and training, many recent works have demonstrated certain flaws in CBMs which warrant an increased caution in their widespread applications. For exmaple, Margeloiu et al.~\cite{margeloiu2021concept} demonstrated that computing the saliency maps with respect to a single concept does not capture the position of that concept in the image itself. Similarly,  Mahinpei et al.~\cite{mahinpei2021promises} demonstrate that CBMs suffer from ``information leakage'' where more than necessary information is encoded in a concept - making them adulterated with non-relevant noise resulting in unreliable downward predictions.


In this paper, we aim to study the security vulnerability and robustness of concept-based models to carefully crafted malicious attacks, where an adversary with a malevolent intent aims to introduce perturbations to clean sample image and modify it in an adversarial manner to manipulate the concepts predicted by the model. Specifically, we first demonstrate how concepts learned by a concept-based model can be manipulated by introducing adversarially generated perturbations in input samples. The goal of attacker is to effectively manipulate concepts \textbf{without} changing the final model predictions. We will study different concept attacks - concept erasure, concept introduction and concept confounding - all of which disrupt the concept set predicted by a well trained concept-based model. Note that proposed attacks can be generalized for any concept-based model. We utilize Concept Bottleneck Models~\cite{koh2020concept} in this paper as an example to demonstrate the efficacy of our attacks on one of the most popular concept-based modelling paradigm. To improve trust and reliability of concept-based models, it is important that both concepts and predictions are robust to malicious attacks. Instilling trust in predictions are a well researched problem in adversarial literature. However, the robustness of concepts is an open question. In our paper, we focus on analyzing robustness of concepts without changing model predictions. This critical difference creates important distinction between our proposed attacks and standard adversarial attacks where goal of attacker is to change the prediction label. 

As shown in Figure~\ref{fig:example-attack-oai}, an attacker can easily introduce non-relevant concepts without changing the prediction label. These non-relevant concepts can cause misinterpretations as shown in Figure~\ref{fig:example-attack-oai} - 
the concept ``osteophytes femur lateral''- which quantifies amount of growth of bone spurs in the upper bone has been maliciously changed to a very high value. This disruption, especially in high security settings,  can prompt remedial actions - like expensive oral medicines or even surgery to fix a ``supposed'' problem even if it does not exist. The fact that such concepts can be manipulated without any perceptible change in the appearance of the input sample and its final prediction essentially defeats the utility of concept-based models in critical applications as these attacks can be very hard to detect. To alleviate this, in addition to studying attacks, we  also propose a general adversarial training-based defense mechanism to improve the robustness of the learning models against the proposed attacks on concepts. We conduct comprehensive experiments on different datasets of varying risk levels - ConceptMNIST, CUB and OAI, and the derived experimental results demonstrate the efficacy of both our attacks and defense.

\section{Related Work}


\textbf{Related work on concept-level explanations}. To incorporate a broader perspective on model decision making in sensitive applications such as medical diagnosis or financial forecasting \cite{SuoBigData20,Xun2020CorrelationNF}, concept attribution methods have been proposed. These methods provide a higher level abstract notion of explanations by aligning model explanations with human-understandable concepts improving overall reliability. 
Several popular methods which automatically learn concepts are detailed \cite{kim2018interpretability,ghorbani2019towards,yeh2020completeness,wu2020towards,goyal2019explaining}. On the other hand providing concept priors have been utilized to align model concepts with human understandable concepts \cite{zhou2018interpretable,murty2020expbert,chen2019explaining}.

\textbf{Related work on concept bottleneck models (CBMs)}. Concept bottleneck models were initially limited to specific use-cases. 
More recently, the applications of such bottleneck models was generalized in the recent work \cite{koh2020concept} which postulated that any prediction model architecture can be transformed into a CBM by simply resizing any intermediate layer to represent a human-understandable concept representation. Similar work on utilizing and improving CBMs for various downstream tasks include \cite{sawada2022concept,jeyakumar2021automatic,pittino2021hierarchical,bahadori2020debiasing}.

\textbf{Related work on robustness of interpretations}. Although explanations have enabled deeper understanding of DNNs, there are concerns regarding their robustness. \cite{ghorbani2019interpretation} showed that explanations can easily be misled by introducing imperceptible noise in the input image. Several other works have highlighted similar problems on vision, natural language and reinforcement learning such as \cite{adebayo2018sanity,dombrowski2019explanations,slack2020fooling,kindermans2019reliability,sinha2021perturbing,huai2020malicious}. Similarly, concept explanation methods are also fragile to small perturbations to input samples~\cite{brown2021brittle}. Such concerns regarding fragility of model explanations have prompted related research in improving robustness of explanation methods. For example, \cite{levine2019certifiably,lakkaraju2020robust,mangla2020saliency} proposed learning more robust feature attributions while \cite{alvarez2018towards,soni2020adversarial,huai2022towards} try to learn more robust concepts. However, existing defense methods \cite{levine2019certifiably,lakkaraju2020robust,mangla2020saliency} on improving the robustness of the feature-level model explanations cannot be directly adopted here. The reason is that they only focus on the post-hoc interpretations, while we work on the intrinsic concept-based interpretable network.


\section{Methodology}
This section investigates vulnerability of CBMs to malicious attacks. We first introduce details of proposed attack strategies. Subsequently, we propose a defense mechanism \textbf{Robust Concept Learning (RCL)}, based on which we train robust models to prevent malicious attacks. Even though our experiments are conducted on CBMs, attacks are general enough and can be used to attack any concept-based model.

\subsection{Malicious Attacks against CBMs}
In this section, we propose a general optimization framework for designing malicious attacks. Here, we use $K$ and $T$ to denote the number of class labels and the number of concepts, respectively. In a CBM model, we are given a set of training samples $\{(x_i,y_i,c_i)\}_{i=1}^{N}$, where $x_{i} \in \mathbb{R}^{D}$ denotes the $i$-th training sample, $y_{i} \in \{1,\cdots,K\}$ is the target classification label for sample $x_i$, and $c_{i} \in \mathbb{R}^{T}$ is a vector of $T$ concepts. CBMs usually consider two components, i.e., the concept component $g(\cdot)$ and the prediction component $f(\cdot)$. Specifically, CBMs consider the form $f(g(x))$, where $g: \mathbb{R}^{D} \rightarrow \mathbb{R}^{T}$ maps an input $x$ into the concept space, and $f: \mathbb{R}^{T} \rightarrow \mathbb{R}^{K}$ maps concepts into the final prediction. CBMs define task accuracy as how accurately $f(x)$ predicts label $y$, and concept accuracy as how accurately $g(x)$ predicts concept $c$. For sample $x$, we use $\mathcal{U}(x;f,g)$ to denote its concept-based explanations generated by $g$ to explain the predicted classification label (i.e., $argmaxf(x)$). Let $G(\mathcal{U}(x;f,g),\mathcal{U}(x+\delta;f,g))$ denote the attacker's goal of maximizing the difference between the generated concept-based explanations before and after the attacks. In order to achieve the attacker's attacking goal, we propose the following framework:
\begin{align*}
\label{eq:maxObjective}
&\max_{||\delta||_{\infty} \leq \epsilon_{thresh}} \quad G(\mathcal{U}(x;f,g),\mathcal{U}(x+\delta;f,g)) \quad  \\ \notag
&\text{s.t.}\quad argmax f(x+\delta)=argmaxf(x), \quad  x + \delta \in [0,1]^{D},
\end{align*}
\vskip -2pt
\noindent where $\delta$ denotes the adversarial perturbation, $\epsilon_{thresh}$ controls the magnitude of the whole adversarial perturbations, and $\mathcal{U}(x+\delta;f,g)$ is the generated concept-based explanations to interpret the predicted class label for the crafted adversarial sample $x+\delta$. Objective function is used to maximize the difference of generated concepts before and after the attacks. The first constraint is enforced to make sure that predictions of sample $x$ is identical before and after the attack. The second constraint guarantees that generated perturbation is imperceptible so it cannot be easily detected. $\textit{l}_{\infty}$ norm is most commonly used when considering imperceptible perturbations and measures the feature with the largest amount perturbation, regardless of number of other features that have been maliciously modified.  By solving the above optimization problem, the attacker can find an optimal perturbation that can maximize the attacker's goals. Depending on how to define the attacker's goal, we categorize three different types of attacking, which are given as following:
\begin{itemize}
    \item \textbf{Erasure:} Concept erasure attack seeks to subtly delete a particular concept without changing the class prediction result. The gap in perception and absence of concepts would be puzzling to an analyzer and very difficult to detect, especially in datasets where every image of the same class does not have the same concepts - while still seemingly giving the same final prediction. Note that in CBMs, the importance score of the $j$-th concept for sample $x$ is calculated as $g_{j}(x)$. In practice, for CBMs, we usually have a pre-defined threshold $\gamma$ that is used to determine whether a concept is a relevant concept. Specifically, for sample $x$, the $j$-th concept is a relevant concept if {$g_{j}(x)-\gamma \geq 0$}. Let $S_{x,Rev}$ denote the set of the targeted initially relevant concepts. In order to remove the presence of an initially relevant concept, the attacker's goal is defined as follows,
    \begin{align}
        & \sum_{j \in S_{x,Rev}} (\mathbb{I} [\gamma -  g_{j}(x + \delta)] - \mathbb{I}[\gamma -  g_{j}(x)]),
    \end{align}
    where $\mathbb{I}[\cdot]$ is the indicator function, $\delta$ denotes the crafted adversarial perturbation, and $\gamma$ is the given threshold. Note that for the $j$-th initially relevant concept, we have $g_{j}(x)-\gamma > 0$, which means that $\mathbb{I}[\gamma -  g_{j}(x)]=0$. The attacker aims to craft the adversarial perturbation $\delta$ such that this $j$-th concept becomes the non-relevant concept, i.e., $\gamma >  g_{j}(x + \delta)$. In other words, the attack is successful if and only if $\mathbb{I} [\gamma -  g_{j}(x + \delta)]=(\mathbb{I} [\gamma -  g_{j}(x + \delta)]- \mathbb{I}[\gamma -  g_{j}(x)])=1$, where $\mathbb{I}[\gamma -  g_{j}(x)]=0$. The above objective is used to maximize the attacker's goal by reducing the importance score of these initially relevant concepts such that their importance scores are less than the threshold $\gamma$. By solving the above objective, the attacker can find an optimal perturbation that can remove the presence of initially relevant concepts for sample $x$. 
    

     \item \textbf{Introduction:} Concept introduction attack aims to manipulate the presence of non-relevant concepts without modifying the classification result. This hinders accurate analysis of model's interpretations by providing mixed interpretations. The attacker tries to introduce new non-relevant concepts which were not previously present in the concept set of the original sample. For sample $x$, let $S_{x,Non}$ denote the set of targeted concepts that do not originally present in sample $x$. The attacker's goal of attacking the presence of these targeted initially non-relevant concepts can be formulated as follows,
     \begin{align}
        & \sum_{j \in S_{x,Non}}(\mathbb{I} [g_{j}(x + \delta)-\gamma]- \mathbb{I}[g_{j}(x)-\gamma]),
    \end{align}
     where $\delta$ denotes the perturbation to be optimized. Note that for the $j$-th initially non-relevant concept, if $g_{j}(x + \delta)-\gamma \geq 0$, we can say this initially non-relevant concept becomes the relevant concept after perturbation. The above loss defines attacker's goal - maximizing presence of targeted non-relevant concepts. Specifically, above loss function aims to maximize the attacker's goal by increasing the importance scores of the targeted initially non-relevant concepts such that these targeted concepts' importance scores are larger than the threshold $\gamma$. To achieve his goal of maximizing the presence of the initially non-relevant concepts for sample $x$, attacker can solve above objective to find an optimal perturbation.
 
     \item \textbf{Confounding:} Concept confounding attack attempts to build on top of both erasure and introduction by simultaneously removing relevant concepts and introducing non-relevant concepts. The concept confounding attack is a much more powerful attack than just the concept introduction attack as it also removes concepts while maintaining the same model prediction. This can be especially troublesome as it would defeat any purpose of training models with concept bottlenecks. Let $S_{x,Rev}$ and $S_{x,Non}$ denote index set of the targeted initially relevant concepts and the set of the targeted initially non-relevant concepts, respectively. In this case,  attacker's goal can be mathematically represented as follows,
     \begin{align}
        &\sum_{j \in S_{x,Rev}}(\mathbb{I} [\gamma -  g_{j}(x + \delta)]- \mathbb{I}[\gamma -  g_{j}(x)]) \\
        &+ \sum_{j \in S_{x,Non}}(\mathbb{I} [g_{j}(x + \delta)-\gamma]- \mathbb{I}[g_{j}(x)-\gamma]), \notag
    \end{align}
    \noindent where $\delta$ denotes the adversarial perturbation to be optimized. The above objective is used to maximize the attacker's goal by decreasing the importance scores of these targeted initially relevant concepts to reduce their presence and increasing these non-relevant concepts' importance scores to introduce their presence.
    
\end{itemize}
The above schemes define attacker's goals from different aspects. The exact attack formulation and the final optimization equation is given by Equation~\ref{eq:final-attack-formulation} in Appendix. Based on above proposed adversarial attacks, we can perform the security vulnerability analysis to understand how motivated attackers can craft malicious examples to mislead CBMs to generate wrong concepts. The magnitude of perturbation reflect the robustness of CBMs to attacks. The smaller the magnitude of the crafted adversarial perturbations is, the less robust the generated concepts are to the adversarial attacks.

\begin{figure*}[t]
    \centering
    \includegraphics[width=0.9\textwidth]{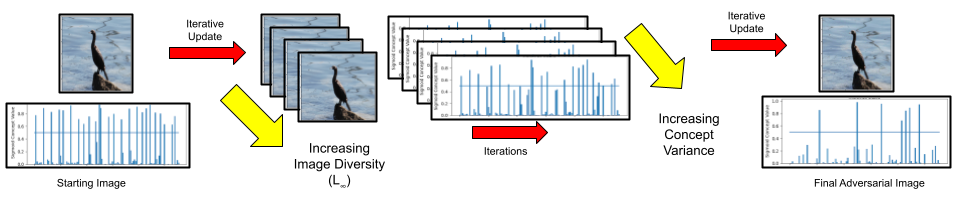}
    \caption{\small \centering Iterative perturbations to generate diverse training images. Corresponding histograms represent presence of concepts across the spectrum. Proposed augmentation generates images that should belong to same concept class but contain wider variance of concepts.} 
    \label{fig:adv-pert}
    \vskip -15pt
\end{figure*}

\subsection{Improving Concept Robustness}
\label{subsec:defense}
Our goal here is to design a defense mechanism which can effectively generate concept-based explanations robust to malicious attacks. Note that in CBMs, we consider bottleneck models of form $f(g(x))$, where $g$ maps an input into the concept space and $f$ maps concepts into a final class prediction. 
Let $\mathcal{L}_{Y}=l(f(g(x_i);y_i))$ and $\mathcal{L}_{C} = \Sigma^{T}_{j}l(g_{j}(x_{i}),c_i^{j})$ denote the classification training loss and the concept training loss over the $i$-th training data, respectively, where $T$ is the total number of concepts and $l$ represents Binary Cross Entropy or Root Mean Square Error loss. 

\textbf{Hybrid training paradigm:} In order to learn the concept component $g$ and the class prediction component $f$, traditional works \cite{koh2020concept,margeloiu2021concept} usually adopt two common ways of learning CBMs - sequential and joint. We discuss both paradigms in brief below:
\begin{itemize}
    \item \textbf{Sequential Training:} Learns the concept model $g$ by minimizing the concept training loss and subsequently learns the class prediction model $f$ by minimizing the classification loss  independently. Mathematically it can be thought of minimizing training objective detailed in Equation~\ref{eq:training-objective} first with $\gamma = 0$ and then subsequently minimizing with $\lambda=0$. As concepts once learned are never updated again during prediction model optimization, the concepts learned are completely independent of the prediction task.
    \item \textbf{Joint Training:} Learns both concept and prediction models ($f$ and $g$) by minimizing both concept and classification loss jointly in an end-to-end manner. Mathematically it can be thought of minimizing the entire training objective Equation~\ref{eq:training-objective} with appropriate values of $\gamma$ and $\lambda$. As concepts and prediction task are learned jointly, concepts learned are not independent of the prediction task as there is some guidance of gradient directions from the prediction part of the model $f$ in the concept model $g$.
\end{itemize}
As demonstrated in Table~\ref{tab:benchmark-results-new} (Appendix), sequential training has lower concept error but worse task performance as compared to joint training (Consistently shown by Figure 2 in \cite{koh2020concept}). Hence, there exists a tradeoff between concept and task loss while using joint or sequential training paradigm. However, as we will demonstrate in Tables~\ref{tab:erasure-results}, \ref{tab:introduction-results} and \ref{tab:confounding-results}, joint training shows higher vulnerability of concepts to malicious attacks - implying that concepts learned during joint training are less robust as compared to those learned in sequential. This behavior is expected - as concepts learned during joint training have higher chances of being spuriously correlated to predictions, making them easier to be maliciously attacked.

To overcome this and achieve a better trade-off between concept robustness and prediction performance, we propose a new hybrid training paradigm by combining the sequential and joint training methods. Specifically, in our proposed hybrid training method, we first freeze the prediction model and only let the concept model learn for the first half of total epochs. Subsequently, we unfreeze the complete model and let training continue for the remainder of epochs with a lower learning rate. Based on this, we formulate training loss as follows, where $(x_i,c_i,y_i)$ is a data point sampled from image set (X), concept set (C), and label set (Y):
\begin{align}
    \label{eq:training-objective}
    \mathcal{L}_{f,g} = & \Sigma_{i}[\gamma * l(f(g(x_{i});y_{i})) 
    + \lambda*\Sigma^T_{j} l(g_j(x_{i});c^{j}_{i}) ],
\end{align}
where the first and second terms represent the task and concept losses for $i$-th training sample with $T$ total number of concepts, respectively. The values of $\gamma \in \{0,1\}$ and $\lambda \in \mathbb{R}$. Using the above loss formulation, the complete model parameters $\theta_{f,g}$  are updated as follows:
\[
\theta_{f,g} = 
\begin{cases}
\theta_{g} - \omega*\nabla_{\theta_{g}}~\mathcal{L}_{f,g}\ \, \, (\gamma=0) & \text{if epoch~$\leq$~N/2,}\\
 \theta_{f,g} - \omega^{'}*\nabla_{\theta_{f,g}}~\mathcal{L}_{f,g} & \text{epoch~$>$~N/2}
\end{cases}
\]
where $\omega$ and $\omega^{'}$ represent learning rates. The above proposed hybrid training method is a two-stage training paradigm. Specifically, during the optimization procedure, for the first half of epochs, we set $\gamma=0$ and $\lambda=1$ and learning rate $\omega$ such that we can first freeze the class prediction model and only train the concept model. Subsequently, in the remaining epochs, we focus on full model training by setting $\gamma$ as $1$ and assigning a pre-defined appropriate weight value to $\lambda$ and a different (smaller) learning rate $\omega$'. The specifics of the training procedures are further detailed in Appendix.

\textbf{Generate diverse training data using adversarial augmentation}.
The essential reason why an attacker can easily introduce malicious perturbations in a sample is the lack of sample diversity in each concept class. The data distribution of each concept can be discrete and highly dispersed. For example, in the CUB (birds) dataset \cite{wah2011caltech} - a sample set containing numerous different types of birds with e.g. `WingColor==Black' concept class, which would still not be enough to cover all possible combinations of birds of different sizes, shapes, etc. Hence, the distribution of `WingColor==Black' concept has huge vacancies in its domain that CBM fails to explore while malicious attacker can easily manipulate. One way to make it difficult for the attacker to exploit such `vacancies' in data distribution (previously unexplored by CBMs) is to augment the training set by injecting diverse training samples which smoothen the concept distribution space. Intuitively, it simulates a weak attacker and generates images that look perceptually similar - but with potentially different concept classes which in turn, significantly enriches the spectrum of concepts existing in the training data. 

\textbf{Robust concept learning (RCL).} 
\noindent We introduce our proposed approach to effectively generate robust concept-based explanations here. Our framework alternates between an inner maximization, where images are iteratively updated with perturbations that increase diversity in concept distribution; and an outer minimization, where model parameters are optimized to find a sweet spot between class prediction, concept accuracy, and concept robustness.
Specifically, in the inner loop, we aim to find a perturbed input $\tilde{x}_i$, such that, its difference from true input $x_i$ is smaller than a budget $\epsilon_{thresh}$ (i.e., $\|x_{i}-\tilde{x}_i\|_{\infty} \leq \epsilon_{thresh}$), while it maximizes the concept divergence loss $l(g(\tilde{x}_i),c_i))$ (i.e. the concept misclassification error) at the same time. The motivation is to generate images that appear identical but are widely diversified in terms of concept distribution. Formally, we iteratively update $\tilde{x_i}$ in the inner loop as follows,

\begin{align}
    \tilde{x_{i}}~\gets~\tilde{x_{i}}+\epsilon*sign(\nabla_{\tilde{x_i}}l(g(\tilde{x}_{i}),c_{i})) 
\end{align}

Figure~\ref{fig:adv-pert} provides an illustration of how the inner loop is effective in generating images with high diversity. We plot the original image as well as intermediate images at each updating step and the ultimate generated perturbed image along with their associated concepts. As can be seen from the concept histogram of each image - concept distributions vary without much perceptual changes in image.

Once the perturbed sample is iteratively generated, in the outer loop, we aim to optimize the model weight such that it achieves a good balance between task classification, concept prediction as well as concept robustness. The updated total loss $\mathcal{L}_{f,g}$ we optimize is,
\begin{align}
    \mathcal{L}_{f,g}~ =\Sigma_i [ \gamma*{\mathcal{L}_{Y}} + \lambda*{\mathcal{L}_{C}} + \alpha*\mathcal{L}_{adv} ]
\end{align}
where $\mathcal{L}_{Y}$ and $\mathcal{L}_{C}$ denote task classification loss and concept prediction loss, respectively. The adversarial loss $\mathcal{L}_{adv}$ is calculated by $\mathcal{L}_{adv}=~l(g(\tilde{x}_{i}),c_{i})$ ($l$ is the same as defined before). 
$\gamma$, $\lambda$, and $\alpha$ are tunable weights. 

To combine the advantages from both joint and sequential models, we adopt a hybrid training paradigm as described previously, in which we disable the training of prediction model and only allow the concept model to be trained in the first half, before unfreezing the prediction model and training the whole model with a lower learning rate in the second half. Our empirical investigation shows this hybrid paradigm outperforms both sequential and joint training alone by a nontrivial margin. Pseudocode of RCL detailed in Algorithm~\ref{algo:adv-training} is moved to appendix due to space limit.

\section{Experimental Study}

 \subsection{Dataset Description}
We test the proposed approaches on the following 3 datasets of varying domains and levels of security and trust required. For a standard classification task such as digit or bird identification, a wrong concept set is not a very concerning outcome - however for a medical diagnosis - a wrong concept set can be catastrophic. For a more comprehensive description of datasets, please refer to Appendix.
\begin{itemize}
    \item \textbf{ConceptMNIST (C-MNIST):} We augment the original MNIST dataset by constructing concepts of each image by including 2 physical characteristics of numbers in the image along with 10 standard non-overlapping concepts representing one hot encodings of the number, resulting in a size 12 concept vector for each image.  \textbf{[Low Risk]}
    
    \item \textbf{CUB:} The Caltech-UCSD Birds-200-2011 dataset \cite{wah2011caltech} consists of photos of 200 classes of birds. Pre-processing of the dataset is performed exactly as \cite{koh2020concept}. Final dataset consists of 112 concepts for each class with concepts representing physical traits of the birds like wing color, beak size, etc. \textbf{[Low Risk]}
    
    \item \textbf{OAI:} The Osteoarthritis Initiative (OAI) dataset \cite{nevitt2006osteoarthritis} consists of X-ray images and clinical data for about 36,000 patients over 4 years of study who pose a risk of knee osteoarthritis. The task is X-ray grading into 4 different risk categories (KLG Score). Each image has 10 medical concepts from X-ray images such as bone spacing. For more comprehensive description, refer to \cite{pierson2019using}. \textbf{[High Risk]}
    
\end{itemize}
\subsection{Benchmarking and Ablation Study}
\vspace{-1pt}
We train CBMs on all 3 datasets using different training strategies - sequential and joint proposed by \cite{koh2020concept} and hybrid as previously discussed with hyperparameters mentioned in Appendix. We train the respective models for CUB and OAI datasets based on the hyperparameters mentioned in \cite{koh2020concept} as well as train hybrid models on both datasets to compare with standard models. In addition, we also train robust models using RCL (Algorithm~\ref{algo:adv-training} (Appendix)) utilizing both joint and hybrid training paradigms. The task errors for C-MNIST and CUB are classification error while for OAI, task error is Root Mean Square Error (RMSE) as the prediction label is a continuous variable. Concept error for C-MNIST and CUB is 0-1 error (binary concepts), while for OAI, concept error is RMSE (concepts are continuous variables).  The benchmark results are reported in Table~\ref{tab:benchmark-results-new}. As expected, performance of hybrid models lie between joint and standard models in task and concept performance. Usage of all 3 training paradigms presents a trade-off between task and concept performance depending on use-case. For eg., in high-risk settings, where concept accuracy is paramount (e.g. medical diagnosis), sequential can be utilized. Whereas tasks where small errors in concepts can be tolerated but prediction performance is important, joint can be utilized. Hybrid paradigm provides a good trade-off between both sequential and joint paradigms.
\vspace{-5pt}

\subsection{Attack Results and Discussion}
\vspace{-2pt}
We report results on a set of 500 randomly chosen samples from the test set for all 3 datasets. We skip all samples which - a) have wrong task prediction label and b) have concept accuracy $\le$ 60\% for binary valued concepts (C-MNIST, CUB) or concept Root Mean Square Error (RMSE) $\ge$ 0.6 for continuous valued concepts (OAI). In all our experiments, we begin by reporting attack success results using standard adversarial attack setting on the joint model (Adv. Attack (Joint)), followed by results for proposed attacks on standard Joint, Sequential and Hybrid models, and finally on joint and hybrid models trained using RCL. As concept scores are not explicitly used during optimization in standard adversarial setting, we expect attack success metrics to be relatively low. Mathematically standard adversarial setting can be formulated as Equation~\ref{eq:final-attack-formulation} (Appendix) with $\beta$ set to 0.

\textbf{Hyperparameter Selection}
All attacks are performed with 2 distinct sets of hyperparameters. The first set of hyperparameters controls properties of attacks - budget ($\epsilon_{thresh}$), number of steps ($N$) and learning rate ($\epsilon$). We refer popular benchmarks\footnote{\url{github.com/MadryLab/mnist_challenge},\url{cifar10_challenge}} for hyperparameter selection decisions. The second set of hyperparameters controls the influence of concepts ($\alpha$) and influence of predictions ($\beta$) to the loss optimized during attacks (Equation~\ref{eq:final-attack-formulation} - Appendix). We defer discussion around hyperparameters to Appendix.


\begin{table}[t]
\centering
\begin{tabular}{c|c|c|c}
\hline
                    & C-MNIST  & CUB & OAI \\
\hline
Adv. Attack(Joint) & $4\pm0$\% & $1\pm0$\%         & $0\pm0$\%                     \\
\hline
Joint      & $67\pm5$\% & $66\pm7$\%         & $62\pm3$\%                     \\
Sequential & $44\pm4$\% & $56\pm4$\%         & $54\pm5$\%                     \\
Hybrid     & $51\pm4$\% & $59\pm6$\%         & $54\pm4$\%                     \\
\hline
RCL-Joint     & $22\pm2$\% & $32\pm 2$\%         & $5\pm2$\%                      \\
\textbf{RCL-Hybrid}    & $18\pm2$\% & \textbf{$23\pm2$\%}         & \textbf{$1\pm0$\%}                      \\    
\hline
\end{tabular}
\vspace{-8pt}
\caption{Attack results on erasure attacks for datasets - C-MNIST, CUB and OAI averaged over 3 different seeds.}
\label{tab:erasure-results}
\vspace{-13pt}
\end{table}







\begin{figure}[h]
    \begin{minipage}[c]{0.45\textwidth}
        \includegraphics[width=\textwidth]{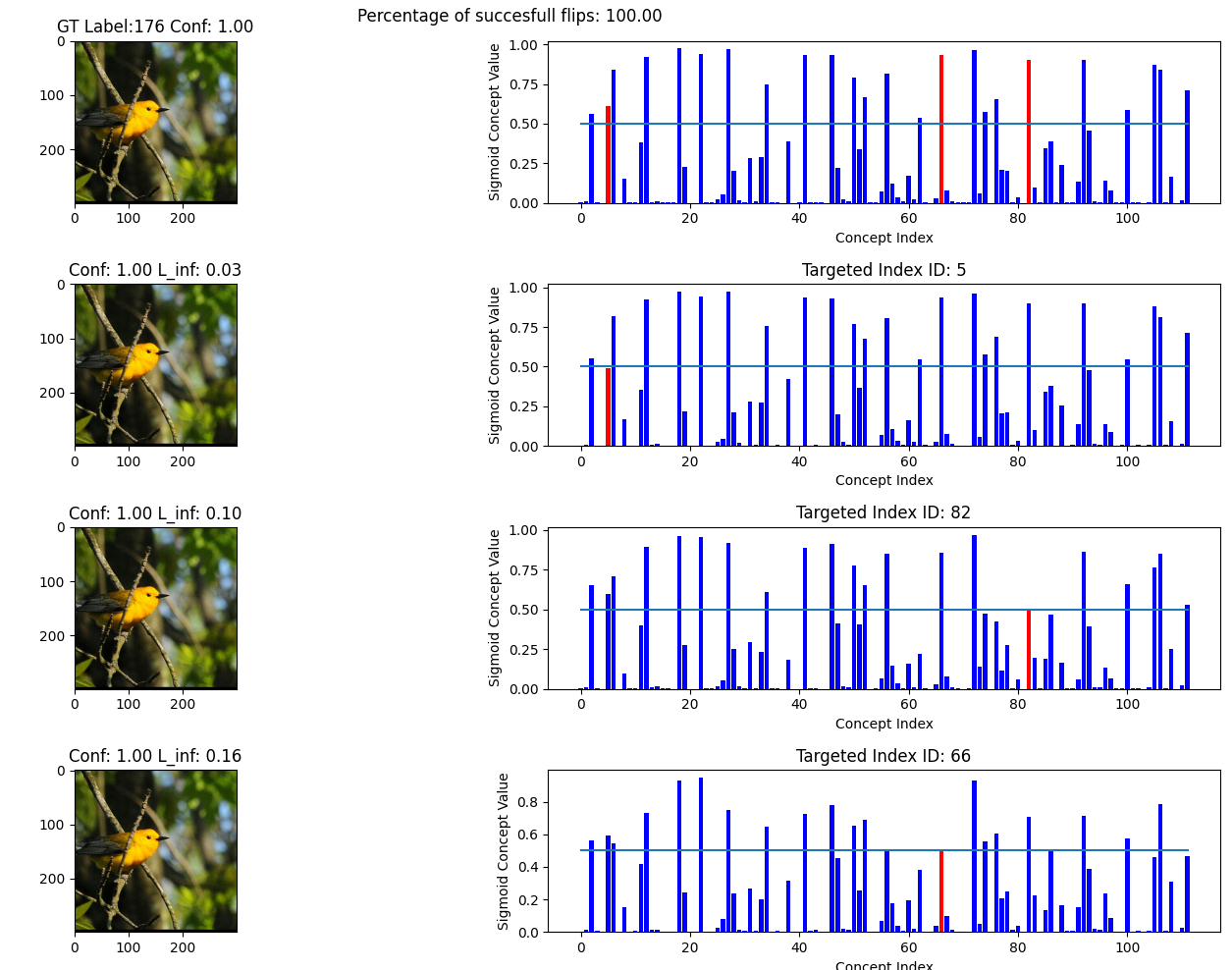} 
        \caption{Top-most: Original image and associated concepts. Following 3 images show final concept set after attack on selected concept (red). Concepts in red previously classified as ``present'' selectively attacked and removed.}
    \label{fig:eg-cub-ers}
    \end{minipage}
    \begin{minipage}[c]{0.45\textwidth}
    \includegraphics[width=\textwidth]{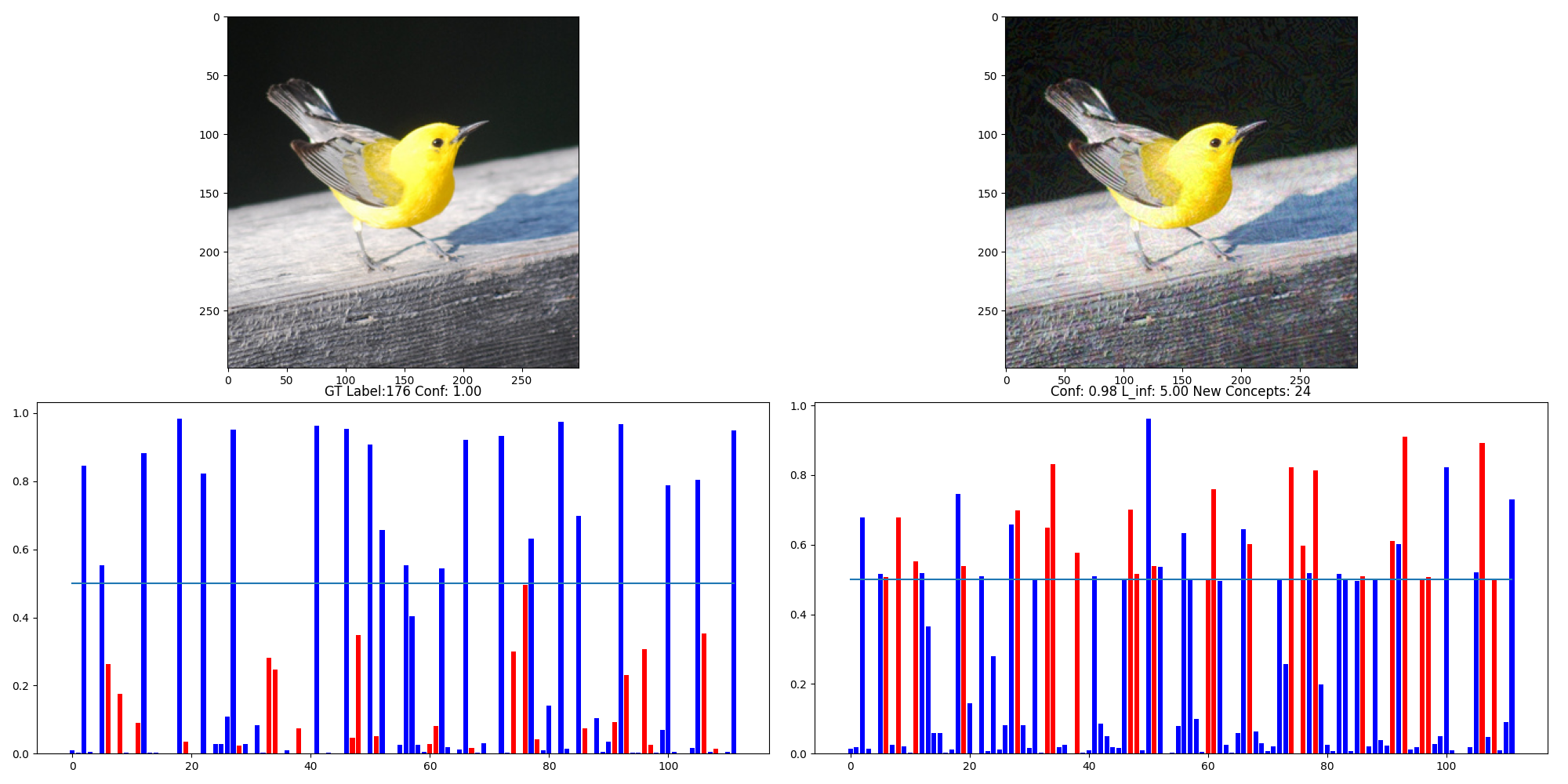}
    \caption{Left: Original image and associated concepts. Right: Concepts in red previously not present in image have been ``introduced'' in perturbed version.}
    \label{fig:eg-cub-intro}
   \end{minipage}
  \vskip -23pt
\end{figure}

\textbf{Results on Erasure Attack}.
As erasure attack attempts to remove or ``flip'' relevant concepts in a particular sample, we run our attack by targeting all possible concepts for each selected sample. For C-MNIST and CUB, we classify a sample as being ``flipped'' if it is no longer classified as being `present' based on sigmoid classification ($\geq$0.5) after the attack. For OAI, we consider a concept as being ``flipped'' if its absolute value changes with more than a pre-defined threshold after attack. In the experiments we set this threshold as 2 (hyperparameter settings - Appendix) which we believe can result in a significant shift in medical diagnosis of knee-pain. Table~\ref{tab:erasure-results} shows the percentage of successful flips for standard adversarial attack, followed by joint, sequential and hybrid models across all 3 datasets. A higher percentage of flipped concepts implies a higher success rate for the attack. We observe about 60\% of concepts are successfully flipped across 3 datasets for joint, sequential and hybrid models with the highest and lowest success rates being on joint and sequential respectively as discussed before. Joint and hybrid models trained using RCL show significantly lower attack success rates of 18\%, 23\% and 1\% on C-MNIST, CUB and OAI - demonstrating RCL's success as a defense. As targeted concept scores are not used during optimization in standard adversarial attack setting, we observe successful flip percentages to be low (4\%, 1\% and 0\% for C-MNIST, CUB and OAI). Figure~\ref{fig:eg-cub-ers} demonstrates attack results on a sample from CUB dataset. 

\begin{table}[t]
\centering
\resizebox{0.49\textwidth}{!}{%
\begin{tabular}{c|cc|cc}
\hline
                          & \multicolumn{2}{c}{C-MNIST}                         & \multicolumn{2}{c}{CUB}                \\
                          \hline
              & \multicolumn{1}{l}{\%Intro.} & \%Ret. & \multicolumn{1}{l}{\%Intro.} & \%Ret. \\
\hline
Adv. Attack(Joint) & $53\pm2$\%                               &    $86\pm2$\%             & $8\pm2$ \%                             &     $77\pm4$\%               \\
\hline
Joint      & $114\pm4$\%                               &    $96\pm2$\%        & $33\pm5$\%                             &    $92\pm3$\%                \\
Sequential   & $71\pm2$\%                              &    $93\pm5$\%       &  $30\pm4$\%                             &     $97\pm2$\%              \\
Hybrid   & $102\pm2$\%                              &    $94\pm2$\%          & $31\pm3$\%                              &     $95\pm2$\%               \\
\hline
RCL-Joint     & $18\pm3$\%                              &     $96\pm2$\%       & $13\pm2$\%                              &      $93\pm3$\%              \\
\textbf{RCL-Hybrid} & \textbf{$13\pm2$}\%                     & \textbf{$96\pm2$\%} & \textbf{$23\pm2$\%}                     & \textbf{$97\pm2$\%}     \\
\hline
\end{tabular}
}
\caption{Attack results on introduction attacks for datasets - C-MNIST and CUB averaged over 3 different seeds. \%Intro denotes the percentage of new concepts introduced wrt. original concept set, while \%Ret denotes the percentage of concepts retained from the original concept set. Note: if more than original number of concepts are introduced, introduction percentage $\geq$100\%.}
\label{tab:introduction-results}
\vspace{-15pt}
\end{table}
\textbf{Results on Introduction Attack}.
As opposed to erasure, introduction attack attempts to introduce non-relevant concepts to concept prediction set of a perturbed sample image. As introduction attack specifically targets non-relevant concepts, this attack is not suitable for data with continuous concept values (e.g. on OAI, all concepts are deemed to be relevant for prediction). We report percentage of new concepts introduced (\%Introduced) in perturbed image before and after attack. Goal of attack here is to introduce previously non-relevant concepts, hence higher value of \%Introduced implies higher success of the attack. In addition, we also report percentage of concepts retained (\%Retained) from the original concept set to ensure no significant change in originally relevant concepts (ideally close to 100\%).
Table~\ref{tab:introduction-results} shows the percentage of non-relevant concepts successfully introduced on standard adversarial setting followed by Joint, Sequential and Hybrid models across CUB and C-MNIST datasets. For CUB, around 33\%, 30\% and 31\% while for C-MNIST, 114\%, 71\% and 102\% concepts are successfully introduced for joint, sequential and hybrid models respectively. Average percentage of relevant concepts retained are relatively high ($\geq$90\%) for all 3 models. As before, models trained with RCL are less susceptible to attack, with introduction percentages around 20\% for both datasets. Similar to erasure, non-relevant concept scores are not explicitly used during optimization in standard adversarial attack, we observe low values of both percentage introduced and retained. Figure~\ref{fig:eg-cub-intro} demonstrates attack results visually on a sample from CUB.

\textbf{Results on Confounding Attack}.
Confounding is a combination of both erasure and introduction attacks. As confounding essentially maximizes the difference between original and perturbed concept sets, we report the Jaccard Similarity index (JSI) for binary concepts (CUB, C-MNIST) and average (Avg-$\Delta$) and minimum (Min-$\Delta$) absolute change  in concept values for continuous concepts (OAI). Lower JSI values indicate a greater difference in concept sets before and after attack, implying higher success of confounding attack. Similarly, higher values of Avg-$\Delta$ implies confounding attack disrupts values for all concepts by a significant amount whereas, high Min-$\Delta$ implies that even minimum concept disruption caused is still relatively large - reducing trust in all concept predictions. Table~\ref{tab:confounding-results} reports JSI on CUB and C-MNIST for joint, sequential and hybrid models across all 3 datasets. We observe relatively low values of JSI (around 0.2 for CUB and 0.4 for C-MNIST respectively) showcasing the success of proposed attack. We also observe models trained using RCL demonstrate relatively higher JSI values of around 0.5 for both datasets, thus making them less susceptible attack. Similarly, for OAI dataset, RCL models demonstrate much better robustness against confounding attacks with average absolute change 2 orders of magnitude less than their standard counterparts (0.0019 vs 0.35)  - which further validates success of RCL as a defense mechanism. As before, adversarial attack's  JSI is relatively high as none of relevant and non-relevant concept scores are utilized during optimization (details - Appendix).

\begin{table}[t]
\resizebox{0.49\textwidth}{!}{%
\begin{tabular}{c|c|c|cc}
\hline
                          & C-MNIST     & CUB & \multicolumn{2}{c}{OAI}                                   \\
                          \hline
                          & Jaccard Sim & Jaccard Sim   & Avg-$\Delta$ & Min-$\Delta$ \\
\hline
Adv. Attack(Joint)          &        $0.61\pm0.04$    &      $0.51\pm0.03$     &            $0.21\pm0.03$                &           $0.03\pm0.006$                  \\                 
\hline
Joint               &       $0.38\pm0.03$    &       $0.20\pm0.01$      &                     $0.57\pm0.03$       &           $0.13\pm0.001$                  \\
Sequential             &       $0.44\pm0.02$  &     $0.23\pm0.02$     &       $1.06\pm0.05$                      &          $0.35\pm0.001$                   \\
Hybrid                 &       $0.41\pm0.04$  &     $0.25\pm0.03$ &        $0.7\pm0.03$                     &           $0.21\pm0.001$                  \\
\hline
RCL-Joint         &     $0.52\pm0.03$   &     $0.49\pm0.05$   &         $0.0058\pm$1e$-$4                    &      3.1$e$-4$\pm$1e$-$6              \\
\textbf{RCL-Hybrid}      & $0.55\pm0.05$  &   $0.54\pm0.04$       &                0.0019$\pm$1e$-$4             &    1.8$\mathbf{e}$-5$\pm$1e$-$6                   
\\
\hline
\end{tabular}
}
\caption{Attack results on confounding attacks for datasets - C-MNIST, CUB and OAI avg. over 3 seeds. Jaccard Sim. represents Jaccard Similarity indices (JSI). Lower JSI value implies concept set before and after are more dissimilar.}
\label{tab:confounding-results}
\vskip -10pt
\end{table}

\textbf{Effect of varying attack budget ($\epsilon_{thresh}$)}: We also report additional results with varying attack budgets in the Appendix. As expected, attack success rates increase with increasing value of attack budgets ($\epsilon_{thresh}$). However, with higher $\epsilon_{thresh}$, images start to lose visual imperceptibility implying a trade-off between attack success and budget. 

\vspace{-10pt}
\section{Conclusion}
In this paper, we conducted the first systematic study on malicious attacks against concept bottleneck models (CBMs). Specifically, we first proposed different novel attack methods (e.g., the concept erasure attacks and the concept introduction attacks) to show that current CBMs are vulnerable to adversarial perturbations. To defend such adversarial attacks and enhance the robustness of CBMs against adversarial attacks, we also proposed a generic adversarial training-based defense mechanism. Extensive experimental results on real-world datasets not only show that current CBMs are vulnerable to malicious perturbations, but also demonstrate the effectiveness of the proposed defense mechanism.

\newpage
\section{Acknowledgments}
This work is supported in part by the US National Science Foundation under grants 2213700, 2217071, 2008208, 1955151. Any opinions, findings, and conclusions or recommendations expressed in this material are those of the author(s) and do not necessarily reflect the views of the National Science Foundation.
\bibliography{ref}
\clearpage
\newpage

\appendix
\section{Appendix}
\label{sec:appendix}
\subsection{Dataset Description}
\begin{figure*}[h]
    \centering
    \includegraphics[width=0.7\textwidth]{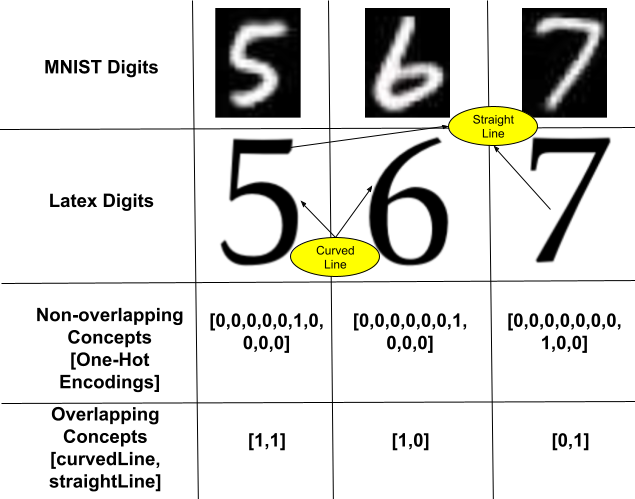}
    \caption{Non-overlapping and overlapping concept construction considered during construction of ConceptMNIST dataset.}
    \label{fig:conceptmnist-construct}
\end{figure*}

We test the proposed attack approaches on the following three different datasets of varying domains and levels of security and trust required. For a standard classification task such as digit or bird identification, a wrong concept set is not a very concerning outcome - however for a medical diagnosis - a wrong concept set can be catastrophic.
\begin{itemize}
    \item \textbf{ConceptMNIST:} We augment the original MNIST dataset which already consists of images of various handwritten numbers from 0-9 along with their labels with carefully constructed, human understandable concepts to emulate real-world image recognition tasks. We propose to augment the dataset with a combination of overlapping and non-overlapping concepts as detailed below:
    \begin{enumerate}
        \item Non-Overlapping Concepts: Certain real-world concepts are shared by samples belonging to only a specific class instance - for example, in medical diagnosis of cancer, the presence of a specific malicious cell. We attempt to emulate such concepts by creating a vector of concepts which are specific to only a class of digits in MNIST. The easiest way to perform this is to one-hot encode labels for each handwritten digit to form a vector of size 10. For example for the number 4, the vector would be represented by [0,0,0,0,1,0,0,0,0,0] where each entry corresponds to a binary concept we call- ``isNum{$_i$}''. Note that no 2 digit classes can share the same vector representation.
        \item Overlapping Concepts: Next, we emulate real-world concepts which are shared by samples from multiple classes - for example, in the CUB bird identification dataset, the color of the bird being black is shared by multiple bird classes. For MNIST, we consider 2 physical characteristics of numbers in the image - presence of a curved line and a presence of a straight line in their \LaTeX visualizations (the way they are printed in \LaTeX typeset formatting). We call these concepts as [``CurvedLine:present'' , ``StraightLine:present''] and are respectively set to 1 if the number can be constructed by only straight lines or curved lines in the standard \LaTeX typeset format. For example, the number 6 has no straight lines but only curved lines - hence is represented as [1,0] while the number 5 has both straight and curved lines and hence represented as [1,1]. Meanwhile, the number 7 has only straight lines so can be represented as [0,1]. Note that multiple digits can share similar representations, for example both numbers 0 and 6 will be represented by [1,0].
    \end{enumerate}
      Both the ``isNum{$_i$}'' and [``CurvedLine:present'' , ``StraightLine:present''] as described above are concatenated - resulting in a size 12 concept vector for each image. Note that this construction of concepts is completely up to user's subjective convenience and can be modified according to what the user thinks best interprets a prediction result. For a visual description of concept annotations, refer to Figure~\ref{fig:conceptmnist-construct}.

    \item \textbf{CUB:} The Caltech-UCSD Birds-200-2011 dataset \cite{wah2011caltech} consists of  11,788 photos of 200 different classes of birds. The task is to classify each photo in one of the categories. The original dataset consists of 312 binary concepts representing physical traits. The dataset is processed exactly as described in \cite{koh2020concept} - final dataset consists of 112 annotated concepts for each class of birds with each concept representing various physical traits of birds like wing color, beak size, etc. It is ensured all birds of same class contain same concept traits.
    
    \item \textbf{OAI:} Osteoarthritis Initiative (OAI) dataset \cite{nevitt2006osteoarthritis} consists of X-ray images and clinical data for about 36,000 patients over 4 years of study who pose a risk of knee osteoarthritis. The task is to perform X-ray grading and classify each image into 4 different risk categories corresponding to Kellgren-Lawrence grade (KLG) which measures the severity of osteoarthritis, with higher numbers implying higher severity. Each image is annotated with 10 concepts which include medical diagnosis from X-ray images such as bone spacing. For a more comprehensive dataset details and processing description, refer to \cite{koh2020concept} and \cite{pierson2019using}. Note that OAI dataset is not publicly available and requires special permissions as detailed in \cite{pierson2019using}.
\end{itemize}

\subsection{Model Training Objectives}
\label{subsec:model-training}
\cite{koh2020concept} proposes 3 different training strategies for Concept Bottleneck Models (CBMs), we only consider joint and sequential model training strategy for all 3 datasets. As \cite{koh2020concept} have demonstrated, joint model training mostly outperforms sequential training approach on task errors and has comparable performance on the concept errors. However, as already discussed previously, both sequential and joint model training have respective advantages with joint offering higher prediction accuracies while sequential offering more robust concept learning. In addition, joint model training is more time-efficient due to being end-to-end in nature and is more flexible to downstream tasks.
Although \cite{koh2020concept} utilizes 3 different training objectives (Section 3, \cite{koh2020concept}),  we formulate them in a unified, general purpose objective function for all types of model training.
\begin{equation}
    \label{eq:training-joint-objective}
    \mathcal{L}_{f,g} = \Sigma_{i}[\gamma*l(f(g(x_{i});y_{i}) + \lambda*\Sigma^T_{j}l(g_j(x_{i});c^j_{i}) ] 
\end{equation}

\begin{itemize}
    \item Sequential Objective: We use Equation~\ref{eq:training-joint-objective} with $\gamma=0$ to first train the model $g$ and subsequently train set $\gamma=1$ and $\lambda=0$ for training the model $f$.
    \item Joint Objective: Equation~\ref{eq:training-joint-objective} can be directly used with appropriate weights for $\gamma$ and $\lambda$ for training both models $f$ and $g$ in tandem.
    \item Hybrid Objective: We use Equation~\ref{eq:training-joint-objective} with $\gamma=0$, to first train the model $g$ and subsequently train set $\gamma=1$ and $\lambda$ to appropriate concept weight value for training both models $f$ and $g$ in tandem.
\end{itemize}

As discussed before, for datasets CUB and OAI, we set the concept weights $\gamma$ to 0.001 and 1 respectively (as reported in \cite{koh2020concept}), while for ConceptMNIST we set $\gamma$ to 0.5 for all subsequent experiments. The learning rates $\omega$ for C-MNIST is 0.01 and $\omega^{'}$ is set as 0.005. For CUB, we follow \cite{koh2017understanding} with $\omega$ at 0.01 and $\omega^{'}$ 0.005  while for OAI $\omega$ and $\omega^{'}$ is set as 0.0008 and 0.0005 respectively. We use SGD momentum as the optimizer and follow tuning rules in \cite{sun22hyperparameters} to set the hyperparameters of the optimizer.

\begin{table}[t]
\centering
\scalebox{0.8}{
\begin{tabular}{c|cc|cc|cc}
\hline
\multirow{3}{*}{}   & \multicolumn{6}{c}{Dataset}                                                           \\
\hline
                    & \multicolumn{2}{c|}{C-MNIST} & \multicolumn{2}{c|}{CUB} & \multicolumn{2}{c}{OAI} \\
                    \hline
                    & Task           & Concept          & Task      & Concept     & Task      & Concept     \\
                    \hline
Joint      &        0.01        &      0.03            &    0.19       &     0.13        &     0.38      &       0.58\\

Sequential &       0.02         &      0.03            &   0.24    &         0.07    &     0.41      &       0.56      \\

Hybrid     &           0.01     &      0.03            &   0.21        &      0.11       &     0.40      &      0.58       \\
\hline

RCL-Joint     &    0.02            &      0.03            &     0.27      &     0.19        &     0.44      &     0.55       \\

RCL-Hybrid    &        0.02        &      0.03            &   0.28        &   0.18          &    0.45       &   0.53    \\
\hline
\end{tabular}
}
\vspace{-8pt}
\caption{Benchmark and replication Task and Concept Errors for 3 datasets over 5 model training paradigms. Models are trained using Joint and Sequential approaches as proposed by \cite{koh2020concept} and hybrid approach as discussed in the Methodology section. Models RCL-Joint and RCL-Hybrid are trained utilizing RCL with joint and hybrid training strategies respectively. Numbers reported are averaged over 3 different seeds with standard deviation under 3\%.}
\label{tab:benchmark-results-new}
\end{table}

\subsection{Model Architectures}
All CBMs consist of two distinct networks - \textbf{$f$} and \textbf{$g$}. We utilize similar model architectures proposed in \cite{koh2020concept} for CUB and OAI datasets. The network $g$ maps inputs in $\mathbb{R}$ to concept space while the network $f$ maps concepts to the output space.
\begin{itemize}
    \item \textbf{ConceptMNIST:} The network $g$ consists of 2 convolutional layers with 32 channels each, along with a maxpool layer in between followed by a fully connected layer. The network $f$ consists of a fully connected layer. The fine-tuning is done end-to-end with the weight 0.1 given to the concept loss. We train the model with batch size set to 64, learning rate set to 1e-4 for a total of 20 epochs. 
    \item \textbf{CUB:} The network $g$ consists of a Inception V3 pre-trained on Imagenet and the network $f$ consists of one layer MLP. The fine-tuning is done end-to-end with the weight 0.001 given to the concept loss. The rest of the hyperparameters are unchanged from \cite{koh2020concept}.
    \item \textbf{OAI:} For the OAI dataset, we utilize the architecture provided in \cite{koh2020concept} which consists of a ResNet-18 with last 12 layers unfrozen. The prediction model consists of a 3 layer MLP with 50 neurons each in the first 2 layers while the last layer provides a regression score. The training hyperparameters are used unchanged.
\end{itemize}
The final benchmarking task and concept losses are reported in the main paper. As can be seen both models trained using standard objective and with robust concept learning with adversarial training obtain similar task and concept errors.






\subsection{Attack Formulation}
\label{sec:attack-formulation}
Recall that our attack framework's objective follows the form detailed in Equation~\ref{eq:maxObjective-appendix}.
\begin{align}
\label{eq:maxObjective-appendix}
&\max_{||\delta||_{\infty} \leq \epsilon_{thresh}} \quad G(\mathcal{U}(x;f,g),\mathcal{U}(x+\delta;f,g)) \quad  \\ \notag
&\text{s.t.}\quad argmax f(x+\delta)=argmaxf(x), \quad  x + \delta \in [0,1]^{D},
\end{align}
The goal of the objective function is to maximize the difference of the generated concept-based explanations before and after the performed adversarial attacks. However importantly, the objective function follows 2 constraints - the first which ensures that model predictions for both the original and attacked samples remain identical, and second that the  generated adversarial perturbation is imperceptible. 

The difference between original and attacked sample's concepts which needs to be maximized, denoted as $\mathcal{D}$ differs for various attacks which are detailed below.
\begin{itemize}
    \item For erasure attack, we attempt to minimize the concept score of a selected target concept, hence the selected set contains only the target  concept score denoted by $C_{S_{target}}$ such that:
    \begin{align}
    \notag
            C_{S_{target}} = &\{s_{target}  | s_{target} \in \mathbb{I}(\sigma(\hat{g}(x_i)){\}}\\ \notag 
            & \mathcal{D}(x_i) = \Sigma~(C_{S_{target}})
    \end{align}
    \item For introduction attack, we attempt to maximize the concept scores of concepts which are not-relevant to the initial model predictions. Therefore, we include all non-relevant concept scores in the selected set denoted by $C_{S_{non-rel}}$ such that:   
    \begin{align}
    \notag
           C_{S_{non-rel}}& = \{s_i | s_i \notin \mathbb{I}(\sigma(\hat{g}(x_i)) \}\\ \notag
            &\mathcal{D}(x_i) = \Sigma(C_{S_{non-rel}}) 
    \end{align}
    \item For confounding attack, we simultaneously maximize concept scores of all relevant concepts and minimize concept scores of non-relevant concepts. The final objective is a weighted sum of relevant and non-relevant concept sets. Therefore, we include all relevant and non-relevant concept scores in the selected set denoted by $C_{S_{rel}}$ and $C_{S_{non-rel}}$ such that:
    \begin{align}
    \notag
             &C_{S_{rel}} = \{s_i | s_i \in \mathbb{I}(\sigma(\hat{g}(x_i)) \}\\ \notag
           &C_{S_{non-rel}} = \{s_i | s_i \notin \mathbb{I}(\sigma(\hat{g}(x_i)) \}\\ \notag
            &\mathcal{D}(x_i) = \Sigma(C_{S_{non-rel}}) + \frac{\gamma}{\beta}*\Sigma(C_{S_{rel}})
    \end{align}
\end{itemize}

However, it is not enough to maximize only $\mathcal{D}$, without considering its effect on model prediction. Note that the first metric specifically ensures model predictions should remain identical after attack. To this effect, we propose to maximize the prediction score of the original sample as output by the prediction model $\hat{f}$. We capture this information in $\mathcal{P}$ which would also be maximized in conjunction with $\mathcal{D}$ to ensure model predictions remain identical.

We have 2 different types of model outputs - standard softmax label-wise outputs for CUB and ConceptMNIST datasets, and a single regressed value for OAI dataset. The different forms of $\mathcal{P}$ are detailed below:
\begin{itemize}
    \item Label-wise outputs (like CUB, ConceptMNIST):
    \begin{align}
        \mathcal{P}(x_i) = -\|\hat{f}(\hat{g}(x_i))_{y_{gt}}\|_{2}\,\,  \\ \notag
         \text{where $y_{gt}$ is index of predicted label}. 
    \end{align}
    \item Continuous outputs (like OAI):
    \begin{align}
        \mathcal{P}(x_i) = -|(f(g(x)) - y_{gt})|\,\, \\
        \text{where $y_{gt}$ is predicted value}. \notag
    \end{align}
\end{itemize}

The third constraint is ensured by stopping optimization before a pre-defined $\epsilon_{thresh}$ value which is kept small enough to ensure visual imperceptibility.

The final objective function $\mathcal{L}$ is thus a weighted sum of $\mathcal{D}$ and $\mathcal{P}$ to maximize the difference in concept values as well as ensure identical model predictions, which is given as follows:

\begin{align}
    \label{eq:final-attack-formulation}
    \mathcal{L}(x_i) = \alpha*\mathcal{P}(x_i)+\beta*\mathcal{D}(x_i).
\end{align}

Algorithm implementation details can be found in Section~\ref{sec:atck-algo} for erasure, introduction and confounding attacks.

\subsection{Attack Settings}

\noindent \textbf{Erasure}.
We set the $\epsilon$ value to $10^{-4}$. We limit our optimization process to 1000 steps and maximum $L_{\infty}$ value to $\epsilon_{thresh}$ so as to ensure visual imperceptibility of added noise. The $\epsilon_{thresh}$ for CUB, ConceptMNIST and OAI datasets is set to 0.2, 30 and 10 respectively for all attacks. We also report results for increasing $L_{\infty}$ values in Table~\ref{tab:linf-erasure}. For all datasets - $\alpha$ and $\beta$ are both set to 1. 

\noindent \textbf{Introduction}.
As before, we set the $\epsilon$ value to $10^{-4}$. We limit our optimization process to 5000 steps and maximum $L_{\infty}$ value to 0.2, 30, 10 as respectively for CUB, ConceptMNIST and OAI datasets as before. We set $\beta$ as 5 to give higher weight to non-relevant concepts and set $\alpha$ to 1.

\noindent \textbf{Confounding}. Confounding attack is a combination of both erasure and introduction attacks. Again we set the $\epsilon$ value to $10^{-4}$, limit our optimization process to 5000 steps and maximum $L_{\infty}$ value to 0.2, 30 and 10  for CUB, ConceptMNIST and OAI datasets respectively. For CUB and ConceptMNIST datasets, the value for $\beta$ and $\gamma$ is set to 10 and 5 respectively.  For OAI dataset, the value for $\beta$ is set at 0.5.

\subsection{Attack Algorithms}
\label{sec:atck-algo}
This section details the specific algorithms to perform general purpose erasure, introduction and confounding attacks. Algorithm~\ref{algo:erasure} details general form of the Erasure attacks as discussed. Algorithm~\ref{algo:introduction} details a general form of confounding attacks. Note that confounding attack with $\gamma$ set to 0 is identical to introduction attack.

\begin{algorithm}[h]

    \DontPrintSemicolon
    \caption{Algorithm to perform concept erasure attack }
    \label{algo:erasure}
    \KwResult{Adversarially perturbed image with same prediction label and concept set with targeted index absent}
    For each image $x_i$ in test set\;
    $C^{orig}\gets\sigma(\hat{g}(x_i))$\;
     $y_{gt}\gets argmax\; \hat{f}(\hat{g}(x_i))$\;
     $x'_i \gets x_i$\,
      $step\gets 0$\,
      $flips\gets 0$\;
     \While{$argmax\; \hat{f}(\hat{g}(x'_i)) == y_{gt}$ \textbf{and}\\ step\,$\leq$\,N}
    {
     $C_{S_{target}}\gets|\sigma(\hat{g}(x'_i))|_{y_{gt}}$\;
     $\mathcal{L}_{x_i}$ = $\alpha*\|\hat{f}(\hat{g}(x'_i))_{y_{gt}}\|_{2}$\; 
      $+\beta*\Sigma(C_{S_{target}})$\; 
      $x'_i \gets x'_i + \epsilon*sign(\nabla_{x'_i}\mathcal{L}_{x_i})$\;
     $x'_i \gets clamp(x'_i,0,1)$\;
     \If{$L_{\infty}(x'_i)\geq$ $\epsilon_{thresh}$}
     {\textbf{break}}
     \If{$\sigma(C_{idx}^{pred})<0.5$}{$flips\gets flips+1$\;\textbf{continue}}
      $step\gets step+1$}
\end{algorithm}
        \begin{algorithm}[h]
        \DontPrintSemicolon
        \caption{A standard algorithm to perform Introduction and Confounding attacks}
        \label{algo:introduction}
        \KwResult{Adversarially perturbed image with same prediction label and different concept set}
        For each image $x_i$ in test set \;
         $x'_i\gets x_i$\;
         $C^{orig}\gets\sigma(\hat{g}(x_i))$\;
         $S_{non-rel}\gets\mathbb{I}(C^{orig}<0.5)$\;
         $S_{rel}\gets \mathbb{I}(C^{orig}\geq0.5)$\;
         $y_{gt}\gets argmax\; \hat{f}(\hat{g}(x_i))$\;
         $step\gets 0$

        \While{$argmax\; \hat{f}(\hat{g}(x'_i)) == y_{gt}$ \textbf{and} step\,$\leq$\,N}
        {
         $C_{pred}\gets\sigma(\hat{g}(x'_i))$\;
         $\mathcal{L}_{x_i}$ = $\alpha*\|\hat{f}(\hat{g}(x'_i))_{gt}\|_{2}$\; 
          $+\beta*\Sigma(C_{S_{non-rel}})$\; 
         $+\gamma*\Sigma(C_{S_{rel}})$\;
         $x'_i \gets x'_i + \epsilon*sign(\nabla_{x'_i}\mathcal{L}_{x_i}))$\;
         $x'_i \gets clamp(x'_i,0,1)$\;
         \If{$L_{\infty}(x'_i)\geq \epsilon_{thresh}$}
         {\textbf{break}}
        $step\gets step+1$
        }
    \end{algorithm}

\subsection{Evaluation Metrics}
In the following, we describe the evaluation metrics in greater detail. 
\begin{itemize}
    \item \textbf{Percentage of concepts flipped:} Erasure and confounding attacks for binary valued concepts attempt to ``flip'' concepts already present in a sample, i.e., the concept is no longer predicted as being relevant to prediction after the attack. Equation~\ref{eq:eval-metric-flip-bin} details the calculation of percentage of flipped concepts for each attacked sample. The terms $C^{orig}$ and $C^{pert}$ denote the concept sets of the original sample and the sample after attack respectively. The indicator function $\mathbb{I}$ returns 1 if concept $i$ is present in the set, else returns $0$. Note that $C^{pert}$ is generated separately for each target concept.
    \begin{equation}
        \text{\% Flipped} = \frac{1}{|C^{orig}|}\Sigma_{i}^{|C^{orig}|}\|\mathbb{I}(C^{orig}_{i}) - \mathbb{I}(C^{pert}_{i})\|
        \label{eq:eval-metric-flip-bin}
    \end{equation}
    However, the notion of concepts ``present'' does not make sense in datasets with real-valued continuous concept annotations. This is because each value of each concept has a certain degree of physical manifestation in the sample - implying a concept is always present. Hence, we consider a concept removed if the absolute difference between the original concept set and the concept set after attack is greater than a threshold value $\delta_{thresh}$.  Equation~\ref{eq:eval-metric-flip-cont} details the calculation of percentage of flipped concepts for each attacked sample. The indicator function $\mathbb{I}$ returns 1 if the condition is true, else returns 0. Note that $C^{pert}$ is generated separately for each target concept.
    \begin{equation}
        \text{\%Flipped} = \frac{1}{|C^{orig}|}\Sigma_{i}^{|C^{orig}|}\|\mathbb{I}(|C^{orig}_{i} - C^{pert}_{i}|>\delta_{thresh}))\|
        \label{eq:eval-metric-flip-cont}
    \end{equation}

    \item \textbf{Percentage of concepts introduced and retained:} Introduction attacks attempt to introduce concepts which are not already present in the sample's concept prediction and are non-relevant to sample's prediction. To this effect, we report two relevant metrics - percentage of concept introduction which refers to the number of new concepts introduced as compared to the original concept set detailed in Equation~\ref{eq:eval-metric-intro-bin}. As we are only interested in new concept introductions, we also report the concept retention percentage refer to the number of concepts which are shared with the original concept set as detailed in Equation~\ref{eq:eval-metric-ret-bin}.
    \begin{equation}
         \text{\%Concepts Introduced} = \frac{|C^{orig} \cap C^{pert}|}{|C^{orig}|}
         \label{eq:eval-metric-intro-bin}
    \end{equation}
    \begin{equation}
         \text{\%Concepts Retention} = \frac{|C^{orig} \cup C^{pert}|}{|C^{orig}|}
         \label{eq:eval-metric-ret-bin}
    \end{equation}
    
    \item \textbf{Jaccard Similarity Index (JSI):} JSI measures the simmilarity (or dissimilarity) between two different concept sets. We utilize the JSI value to understand the dissimilarity of concepts sets predicted before and after the confounding attacks. Mathematically, it can be thought of the percentage of shared set elements with respect to combined total set elements also called Intersection over Union. In general, the lower the JSI value, the more different the input sets are. The exact mathematical formulation is detailed in Eqution~\ref{eq:eval-metric-jsi}.
    \begin{equation}
         \text{JSI} = \frac{|C^{orig} \cup C^{pert}|}{|C^{orig} \cap C^{pert}|}
         \label{eq:eval-metric-jsi}
    \end{equation}
    
\end{itemize}


        \begin{algorithm}[h]
            \label{algo:adv-training}
            \DontPrintSemicolon
            \caption{Robust Concept Learning (RCL) }
            \SetKwInOut{Output}{Output}
            \Output{Trained models $\hat{f},\hat{g}$ with increased concept robustness}
            \SetKwInOut{Input}{Input}
            \Input{Image Set: $X$, Concept Set: $C$, Label Set: $Y$, Training Corpus:~\{$x_i\in X$;~$c_i\in C$;~$y_i\in Y$\}, Epochs:~$N$, Models:~$f,g$, Learning Rate:~$\omega$, Step~Size~: $\epsilon$, Max~Iterations:~$num$}
            $epoch\gets 0$\; 
            \While{$epoch\leq N$ }
            {
            Sample a batch of training samples: $Z\, = \, \{(x_{i},c_{i},y_{i})| i = 1,2,...,B\}$\;
            $\mathcal{L}_{C}\gets~\Sigma^{T}_{i}{l}(g_j(x_{i}),c^{j}_{i})$ [note: $l$ - Binary Cross Entropy/Mean Square Error Loss]\;
            ${\mathcal{L}_{Y}}\gets~l(f(g(x_{i})),y_{i})$ [note: $l$ - Cross Entropy Loss]\;
            $s\gets0$\;
            $\tilde{x_{i}}\gets x_{i}$\;
            \While{$s\leq~num$}
            {
              $\tilde{x_i}~\gets~\tilde{x_i}+\epsilon*sign(\nabla_{\tilde{x_{i}}}\Sigma^{B}_{i=1}l(g(\tilde{x_i}),c_{i}))$ [note: $l$ - Binary Cross Entropy/Mean Square Error Loss]\;
            }
            $\mathcal{L}_{adv}\gets~l(f(g(\tilde{x_{i}})),y_{i})$\;
            $\mathcal{L}_{f,g}~ = \Sigma^B_{i}[\gamma*{\mathcal{L}_{Y}} + \lambda*{\mathcal{L}_{C}} + \alpha*\mathcal{L}_{adv}$]\;
            \uIf{epoch~$\leq$~N/2}
                {$\theta_{g} \gets \theta_{g} - \omega*\nabla_{\theta_{g}}~\mathcal{L}_{f,g}$\;}
            \uElse
                {$\theta_{f,g} \gets \theta_{f,g} - \omega'*\nabla_{\theta_{f,g}}~\mathcal{L}_{f,g}$\;}
            }
        \end{algorithm}
        \vskip -15pt

\subsection{Robust Concept Learning (RCL)}
We detail the exact implementation of the proposed defense method Robust Concept Learning (RCL) in this subsection. As discussed before, the essential reason behing the lack of robustness of CBMs is the lack of rich concept information during training. To alleviate this problem, we propose adversarial augmentation to further improve the smoothness of the concept space.  Algorithm~\label{algo:adv-training} iteratively generates increasingly diverse images by optimizing the adversarial loss $\mathcal{L}_{adv}$. The final model optimization occurs with a weighted sum of all 3 - prediction, concept and adversarial loss. The values of $\alpha$ are set to 1 for C-MNIST, CUB and OAI datasets. 

\subsection{Additional Results}
We report additional results on standard joint models and hybrid adversarially concept trained with higher concept robustness over all 3 datasets and attacks for varying $L_{\infty}$ values. As previously discussed, the $L_{\infty}$ values are considered to be a direct measure of visual imperceptibility, i.e the lower the value, the more imperceptible adversarial perturbations introduced in an image. As we postulated, any attack is much more effective and malicious if maximum attack success can be obtained with small $L_{\infty}$ values. We demonstrate the evaluation metrics for all 3 attacks in Tables~\ref{tab:linf-erasure}, \ref{tab:linf-intro} and \ref{tab:linf-conf} for all 3 datasets. As evident from all 3 tables, the models trained using the proposed Robust Concept Learning with Adversarial Training algorithm are much more robust to attacks for small $L_{\infty}$ values, thus improving trust and reliability in CBMs. The values for budget - $\epsilon_{thresh}$ are chosen according to standard benchmarks for adversarial attacks e.g. \url{https://github.com/MadryLab/mnist_challenge}. As MNIST has a smaller latent representation the values for $\epsilon_{thresh}$ are relatively high while CUB is a very high dimensional dataset, the relative $\epsilon_{thresh}$ values are low.

\begin{table}[h]
\centering
\resizebox{0.49\textwidth}{!}{
\begin{tabular}{c|ccc|ccc|ccc}
\hline
                 & \multicolumn{3}{c|}{\textbf{C-MNIST}} & \multicolumn{3}{c|}{\textbf{CUB}} & \multicolumn{3}{c}{\textbf{OAI}} \\
                 \hline
                 & \textbf{10}       & \textbf{30}       & \textbf{50}      & \textbf{0.2}    & \textbf{0.4}    & \textbf{0.5}   & \textbf{10}    & \textbf{15}    & \textbf{20}   \\
                 \hline
Joint            &     53      &    67       &  70        &   65     &    99    & 99      &  62      &  71      &   74    \\
RCL-Hybrid &      18     & 22          &     24     &   23     &    31    &     36  &  1      &   1     &    1  \\
\hline
\end{tabular}}
\caption{Erasure attack success percentages for varying $L_{\infty}$ values. The higher the $L_{\infty}$, the lower the visual imperceptibility. As expected attack success percentages increase with increasing $L_{\infty}$. }
\label{tab:linf-erasure}
\end{table}

\subsection{Additional visual results}
\begin{figure*}[h]
    \centering
    \includegraphics[width=\textwidth]{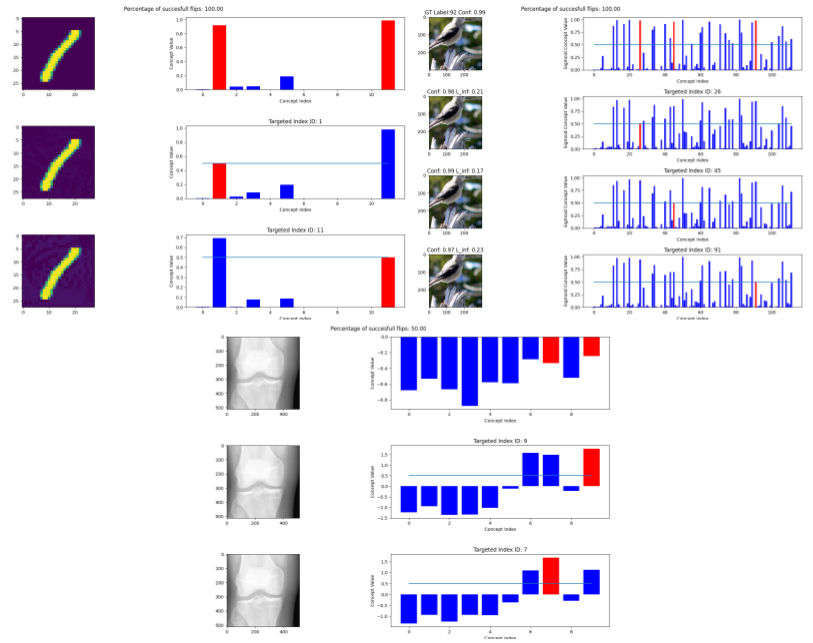}
    \caption{A few examples on the ConceptMNIST (top-left, upper row), CUB (top-right, upper row) and OAI dataset (lower row) demonstrating the erasure attacks. }
    \label{fig:ers-extra}
\end{figure*}

We also present additional visual results for Erasure attacks on the ConceptMNIST (top-left), CUB (top-right) and OAI datasets (bottom-center) in Figure~\ref{fig:ers-extra}. The bars in the first row of each image represent the initial concept scores as predicted by the model. The bars in color red are randomly chosen concepts we attempt to and successfully flip in rows 2 and 3 of each image. (Note: we attempt to flip all present concepts, but demonstrate only 2 random successfully flips). 

\begin{figure*}[h]
    \centering
    \includegraphics[width=\textwidth]{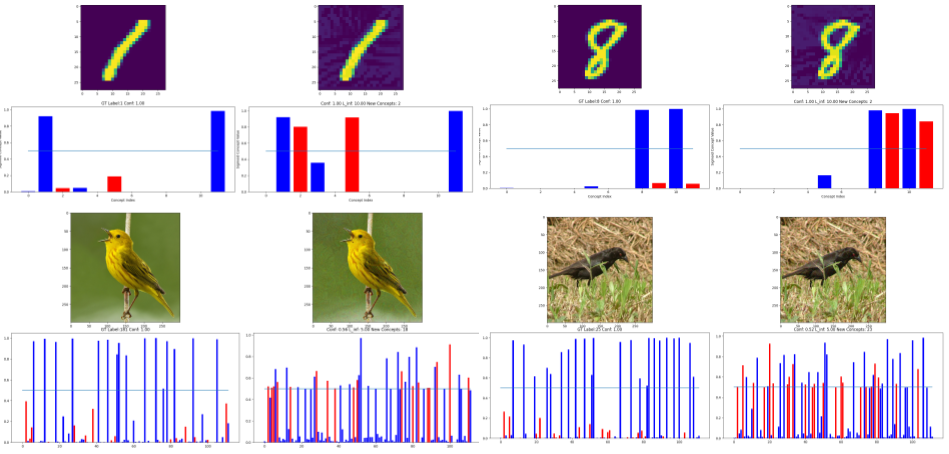}
    \caption{A few examples on the ConceptMNIST (upper row) and CUB dataset (lower row) demonstrating the introduction attacks. }
    \label{fig:intro-extra}
\end{figure*}

Visual results for Introduction attacks on the ConceptMNIST (top-left), CUB (top-right) and OAI datasets (bottom-center) are reported in Figure~\ref{fig:intro-extra}. The first image and the bars below it represent the initial image and the concept scores as predicted by the model, while the second image and bars below it represents the image and concept predictions after the attack. The bars in color red are previously non-relevant concepts which were successfully introduced after attack. The images in the first and last row are for ConceptMNIST and CUB datasets, respectively. 

We release code for data processing, attacks and results on the ConceptMNIST dataset. Upon acceptance, we will also release codes for CUB dataset. As OAI dataset requires special permissions - only limited code will be released for it.

\begin{table*}[t]
\centering
\resizebox{\textwidth}{!}{%
\begin{tabular}{c|cc|cc|cc|cc|cc|cc}
\hline
                       & \multicolumn{6}{c|}{\textbf{C-MNIST}}                              & \multicolumn{6}{c}{\textbf{CUB}}                                            \\
                       \hline
$L_{\infty}$ & \multicolumn{2}{c|}{30} & \multicolumn{2}{c|}{40} & \multicolumn{2}{c|}{50} & \multicolumn{2}{c|}{0.2} & \multicolumn{2}{c|}{0.4} & \multicolumn{2}{c}{0.5} \\
\hline
                       & \%Intro    & \%Ret    & \%Intro    & \%Ret    & \%Intro     & \%Ret    & \%Intro     & \%Ret     & \%Intro     & \%Ret     & \%Intro     & \%Ret     \\
                       \hline
Joint                  &       118     &       96   &      140      &    94      &        145     &    92      &      33       &  92          &   42          &   91        &     47        &     87      \\
RCL-Hybrid       &       13     &      96    &      22      &  95        &     26        &  92        &      7       &    61       &    12         &      97     &      14       &    97      \\
\hline
\end{tabular}}
\caption{Introduction attack introduction and retention rates for varying $L_{\infty}$ values. The higher the $L_{\infty}$, the lower the visual imperceptibility. As expected introduction rates increase with increasing $L_{\infty}$. }
\label{tab:linf-intro}
\end{table*}
\begin{table*}[t]
\centering
\resizebox{\textwidth}{!}{%
\begin{tabular}{c|ccc|cc|cc|cc|cc|cc|cc}
\hline
                             & \multicolumn{3}{c}{\textbf{C-MNIST}}                              & \multicolumn{6}{|c|}{\textbf{CUB}}                                            & \multicolumn{6}{c}{\textbf{OAI}}                                            \\
                             \hline
$L_{\infty}$ & 30 & 40 & 50 & \multicolumn{2}{c|}{0.2} & \multicolumn{2}{c|}{0.4} & \multicolumn{2}{c|}{0.5} & \multicolumn{2}{c|}{10} & \multicolumn{2}{c|}{15} & \multicolumn{2}{c}{20} \\
\hline
                             & JSI       & JSI       & JSI        & $\%$Intro     & $\%$Ret     & $\%$Intro     & $\%$Ret     & $\%$Intro     & $\%$Ret     & Avg        & Min        & Avg        & Min        & Avg        & Min        \\
 \hline
Joint   &  $0.38$  &       $0.32$      &     $0.30$         &  $13$  &   $31$        &     $15$        &   $30$        &     $18$        &  $28$    &      $0.56$      &  $0.13$         &      $0.60$     &  $0.13$          &     $0.62$       &    $0.13$         \\
RCL-Hybrid &      $0.55$        &  $0.52$  &  $0.51$    &  $7$ &  $61$         &     $9$        &      $58$     &    $12$         &    $51$      &   $1.9e$-$3$         &    $1.8e$-$5$        &    $2.6e$-$3$        &     $3.9e$-$5$       &    $5.1$e-$3$        &  $4.1e$-$5$    \\   
\hline
\end{tabular}}
\caption{Confounding attack introduction and retention percentages for varying $L_{\infty}$ values. The higher the $L_{\infty}$, the lower the visual imperceptibility. As expected introduction percentages increase and retention percentages decrease with increasing $L_{\infty}$. }
\label{tab:linf-conf}
\end{table*}


\end{document}